\documentclass[conference]{IEEEtran}
\usepackage[utf8]{inputenc}
\usepackage{graphicx}
\graphicspath{{images/}}
\usepackage[caption=false,font=footnotesize]{subfig}
\usepackage{amsmath,amssymb,amsfonts}
\usepackage{cite}
\usepackage{textcomp}
\usepackage{xcolor}
\usepackage{booktabs}
\usepackage{array}
\usepackage{algorithm}
\usepackage{algpseudocode}
\usepackage[hidelinks]{hyperref} 
\usepackage{xurl}                
\def\BibTeX{{\rm B\kern-.05em{\sc i\kern-.025em b}\kern-.08em
T\kern-.1667em\lower.7ex\hbox{E}\kern-.125emX}}

\begin{document}

\title{WiseOWL: A Methodology for Evaluating Ontological Descriptiveness and Semantic Correctness for Ontology Reuse and Ontology Recommendations\\
\thanks{}
}

\author{\IEEEauthorblockN{1\textsuperscript{st} Aryan Singh Dalal} \IEEEauthorblockA{\textit{Department of Computer Science (of Aff.)} \\ \textit{Kansas State University (of Aff.)}\\ Manhattan, Kansas, USA \\ aryand@ksu.edu or 0000-0003-0720-7306} \and 

\IEEEauthorblockN{2\textsuperscript{nd} Maria Baloch} \IEEEauthorblockA{\textit{Department of Computer Science (of Aff.)} \\ \textit{Kansas State University (of Aff.)}\\ Manhattan, Kansas, USA \\ fmaria@ksu.edu or 0009-0002-6470-9361} \and 

\IEEEauthorblockN{3\textsuperscript{rd} Asiyah Yu Lin} \IEEEauthorblockA{\textit{National Center for Ontological Research (NCOR) (of Aff.)} \\ \textit{University at Buffalo (of Aff.)}\\ Buffalo, NY, USA \\ linikujp@gmail.com or 0000-0003-2620-0345} \and 

\IEEEauthorblockN{4\textsuperscript{th} Anna Maria Masci} \IEEEauthorblockA{\textit{University of Texas MD Anderson Cancer Center (of Aff.)} \\ \textit{The University of Texas System (of Aff.)}\\ Houston, Texas, USA \\ amasci@mdanderson.org or 0000-0003-1940-6740} \and 

\IEEEauthorblockN{5\textsuperscript{th} Kathleen M. Jagodnik} \IEEEauthorblockA{\textit{Department of Computer Science (of Aff.)} \\ \textit{Kansas State University (of Aff.)}\\ Manhattan, Kansas, USA \\ kmjagodnik@ksu.edu or 0000-0002-2755-2097} \and 

\IEEEauthorblockN{6\textsuperscript{th} Hande Küçük McGinty*} \IEEEauthorblockA{\textit{Department of Computer Science (of Aff.)} \\ \textit{Kansas State University (of Aff.)}\\ Manhattan, Kansas, USA \\ hande@ksu.edu or 0000-0002-9025-5538} }

\maketitle

\begin{abstract}
The Semantic Web standardizes concept meaning for humans and machines, enabling machine-operable content and consistent interpretation that improves advanced analytics. Reusing ontologies speeds development and enforces consistency, yet selecting the optimal choice is challenging because authors lack systematic selection criteria and often rely on intuition that is difficult to justify, limiting reuse. To solve this, WiseOWL is proposed, a methodology with scoring and guidance to select ontologies for reuse. It scores four metrics: (i) Well-Described, measuring documentation coverage; (ii) Well-Defined, using state-of-the-art embeddings to assess label–definition alignment; (iii) Connection, capturing structural interconnectedness; and (iv) Hierarchical Breadth, reflecting hierarchical balance. WiseOWL outputs normalized 0–10 scores with actionable feedback. Implemented as a Streamlit app, it ingests OWL format, converts to RDF Turtle, and provides interactive visualizations. Evaluation across six ontologies, including the Plant Ontology (PO),  Gene Ontology (GO), Semanticscience Integrated Ontology (SIO), Food Ontology (FoodON), Dublin Core (DC), and GoodRelations, demonstrates promising effectiveness.
\end{abstract}


\section{Introduction and Motivation}
Many Semantic Web applications rely on ontologies that provide structured vocabularies for knowledge representation. Ontology engineering is a challenging but indispensable step in developing an ontology \cite{dalal2025gliide, singh2025flavonoid}; a call for reuse of existing ontologies is now widespread as a best practice to accelerate ontology development, promote standardization, and share collective knowledge \cite{gomez2006ontological,noy2001ontology,pinto2001methodology,DALAL2026111341,noy2003prompt}. However, suitability for reuse depends on ontology quality. Therefore, efficient methods are necessary to evaluate the quality of ontologies before selection and integration \cite{gangemi2005theoretical,brank2005survey}. 
Adhering to structured methodologies like Methontology \cite{fernandez1997methontology}, DILIGENT \cite{pinto2004diligent}, and NeOn \cite{suarez2010neon}, or employing ontology design patterns \cite{gangemi2009ontology} can ensure consistency in modeling practices. While these methodologies have proven effective in certain contexts, they still face challenges, especially when scaling or handling large knowledge domains that require frequent  \cite{dalal2026echo}. Over the last decade, new strategies have emerged to address these inconsistencies. One such methodology is the Knowledge Acquisition and Representation Methodology (KNARM) \cite{mcginty2018knowledge}, which breaks down the ontology creation process into nine steps. This includes extensive language analysis and interviews with domain experts. There also exist some ontology development workflows, such as OLIVE \cite{zhang2024olive}, that automate the process of creating ontologies and minimize manual effort, thereby decreasing the risk of errors. However, post-hoc evaluation techniques are still necessary to assist users in selecting suitable, proven ontologies among the various candidate ontologies \cite{tartir2005ontoqa,lozano2004ontometric}.

BioPortal Recommender \cite{martinez2017ncbo} assesses ontologies based on how well they align with external factors such as term coverage, popularity, and domain relevance. In contrast, WiseOWL, the methodology that we present here, focuses on the internal qualities of an ontology because all its metrics are derived from analyzing the ontology’s own RDF/OWL graph content --its entities, annotations, relationships, and structure -- without relying on external data. Furthermore, the theoretical underpinnings of ontology evaluation are grounded in structural and conceptual frameworks designed to assess an ontology’s quality. The formal ontology OntoClean \cite{welty2001supporting} introduces key meta-properties—identity, unity, dependence, essence, and rigidity to detect taxonomic inconsistencies. While OntoClean ensures logical precision, it also requires substantial expertise and manual effort from ontology engineers to annotate concepts with appropriate meta-properties, limiting its practical adoption \cite{volker2008aeon}. To address this limitation, the AEON methodology \cite{volker2008aeon} automates the annotation of ontologies with OntoClean meta-properties and supports constraint verification processes. AEON evaluates ontology quality across two main axes: quality criteria (standards adherence to design goals) and aspect criteria (structural soundness of the ontology). Although this domain and task-independent approach provides baseline benchmarks and general quality standards \cite{volker2008aeon}, domain-specific evaluations remain essential for more specialized assessments. Alternatively, Gangemi et al. \cite{gangemi2005theoretical} proposed the oQual ontology quality diagnosis model, derived from the O² semiotic meta-ontology \cite{burton2005semiotic}. This model categorizes ontology quality into three key dimensions: structural (e.g., graph depth, modularity), functional (analogous to precision and recall), and usability (including annotation completeness and provenance) \cite{gangemi2005theoretical}. These dimensions assess an ontology’s formal semantics, domain relevance, and documentation quality \cite{gangemi2005theoretical}. However, as Gangemi’s model is inherently generic, it requires adaptation to specific domains and contexts \cite{wilson2023conceptual}. In contrast, our WiseOWL methodology departs from these theoretical and semiotic approaches by offering a quantitative, data-driven framework that evaluates ontology quality directly through its internal RDF/OWL structure. It minimizes reliance on manual expert intervention and focuses on measurable, reproducible indicators derived from the ontology’s own entities, annotations, relationships, and hierarchy. This practical and automated methodology complements the rigor of earlier frameworks while enhancing scalability, comparability, and accessibility in ontology evaluation.

On the other hand, Metric-based decision support approaches enable users to quantitatively assess how well existing ontologies meet their system requirements. OntoMetrics \cite{lozano2004ontometric} evaluates ontologies by analyzing their structural aspects and uses the Analytic Hierarchy Process (AHP) to perform pairwise comparisons based on criteria such as relevance, completeness, and cost. Practitioners have encountered difficulties implementing the OntoMetrics \cite{lozano2004ontometric} approach, as it can be cumbersome in the case of extensive volume repositories of ontology. In contrast, WiseOWL focuses on the semantic richness and hierarchical depth of an ontology, using natural language processing (NLP) techniques to assess how clearly and meaningfully the concepts are defined within the knowledge base. Most existing post-hoc evaluation frameworks \cite{tartir2005ontoqa} assess syntactic correctness and basic quality indicators such as the coverage of classes or property counts. Moreover, OntoQA \cite{tartir2005ontoqa} not only computes automatic schema measurements (depth, breadth) and Knowledge Base (KB) measurements (distribution of instances), but also computes class measurements (connectivity) and characterizes ontologies quantitatively, making fast comparisons possible without requiring manual specification. The fact that OntoQA is data-driven maximizes efficiency, but it may overlook more conceptual elements that cannot be reflected in numerical measures, such as domain relevance.

Automated tools for diagnosing “pitfalls” further streamline the evaluation process. Zaveri et al. \cite{zaveri2013user} propose a semi-automatic tool that involves the generation and verification of ontology-based schema axioms to identify and correct data quality deficiencies. Their approach grounds quality assessment in real-world usage, revealing extraction flaws and informing ontology selection based on empirical fitness. However, crowd-sourcing introduces variance in assessor expertise and may pose scalability challenges for very large datasets. Consequently, this introduces a fundamental conflict with the Semantic Web’s foundational principle of precise semantic definitions, as established by Berners-Lee, Hendler, and Lassila \cite{berners2023semantic}, who stressed that ontologies must support reliable automated reasoning through unambiguous conceptual structures. Pitfall detection tools (e.g., OOPS! \cite{poveda2014oops}) are online tools developed to support ontology developers in the evaluation and improvement of ontology quality. The OOPS! tool suggests considerable help for pinpointing individual errors, provides importance-based categorization (critical, important, minor), and delivers suggestions for ontology design enhancement \cite{poveda2014oops}. OOPS!’s accessibility and evolving pitfall catalog enable rapid validation, though its rule-based checks may miss domain-specific conceptual errors. To maintain quality post-selection, Roldán-Molina et al. \cite{roldan2021ontology} adapt the Deming cycle for ontologies, integrating ISO/IEC 25000-inspired metrics with graphical dashboards for continuous inspection and defect correction in both manually and automatically generated ontologies. This feedback loop aligns ontology evolution with user needs, yet requires sustained metric monitoring and dashboard maintenance that may strain project resources. Consequently, there exists a need for ontology assessment methodologies that provide a balanced, quantitative measure of documentation, semantic clarity, structural complexity, and hierarchical development \cite{brank2005survey}. Our WiseOWL methodology presented below addresses these gaps.

\section{Methodology}

The WiseOWL pipeline presented here creates a full representation of the assessed ontology’s structural elements upon which metrics are calculated, as discussed below. 

\subsection{Well-Described Metric}

The \textbf{Well-Described} metric evaluates the documentation coverage of an ontology. 
It quantifies the proportion of entities accompanied by at least one piece of 
human-readable descriptive text, rather than measuring annotation volume.

\subsubsection*{Predicate Set Definition}
The calculation is based on a comprehensive set of annotation properties. 
This set includes:
\begin{itemize}
\item Standard properties for labeling and commenting 
  (e.g., \texttt{rdfs:label}, \texttt{rdfs:comment}).
\item Advanced properties for preferred, alternative, and hidden labels 
  from established labeling vocabularies.
\item Properties intended for textual definitions, notes, and scope descriptions.
\item Commonly used properties for synonyms and textual definitions 
  found in scientific and biomedical domains.
\item Any custom property explicitly defined as an 
  \texttt{owl:AnnotationProperty} within the ontology itself.
\end{itemize}

\subsubsection*{Entity Coverage Calculation}
The system identifies a complete set of entities by aggregating all subjects 
declared as \texttt{owl:Class}, \texttt{rdfs:Class}, \texttt{skos:Concept}, 
and all individuals. The metric iterates through each entity.  

For a given entity, it receives a score of $1.0$ if it is the subject of at least 
one triple whose predicate is in the defined set, and $0.0$ otherwise. 
The logic also processes advanced labeling constructs 
(such as \texttt{skos-xl:altLabel}) by checking for their associated literal forms.

\subsubsection*{Score Normalization}
The final score is computed as the mean of these per-entity Boolean scores, scaled to 10:
\[
\text{Well-Described Score} = 10 \times \frac{\sum_{i=1}^{N} s_i}{N}
\]
where $s_i \in \{0, 1\}$ indicates whether entity $i$ has at least one descriptive annotation, 
and $N$ is the total number of entities.  

This result represents the percentage of ontology entities that have at least 
one form of descriptive annotation.

\begin{align*}
\mathcal{E} &= \text{entities (classes } \cup \text{ individuals)} \\
\mathcal{P}_{\text{desc}} &= \text{descriptive annotation predicates} \\
\text{isDescribed}(e) &= \begin{cases}
1, & \text{if } \exists p \in \mathcal{P}_{\text{desc}}, o : (e, p, o) \in \mathcal{G} \\
0, & \text{otherwise}
\end{cases} \\
\text{Score} &= 10 \times \frac{\sum_{e \in \mathcal{E}} \text{isDescribed}(e)}{|\mathcal{E}|}
\end{align*}

\begin{algorithm}[!t]
\caption{Pseudocode for WiseOWL Backend Scoring }
\label{alg:wiseowl}
\footnotesize
\begin{algorithmic}[1]
\State \textbf{Input:} ontology path $P$ \hfill \textbf{Output:} $\{S_{\text{desc}},S_{\text{def}},S_{\text{conn}},S_{\text{hier}}\}$
\State $g \gets$ parse $P$ (auto \textrightarrow\ common RDF/OWL formats)
\State $\mathcal{C} \gets$ classes (OWL/RDFS, SKOS) $\,\cup\,$ subclass endpoints; \ $\mathcal{I} \gets$ individuals typed in $\mathcal{C}$
\State $\mathcal{E} \gets \mathcal{C}\cup\mathcal{I}$;\quad $\mathcal{T} \gets \text{list}(g)$
\State \textbf{Object properties:} declared \texttt{owl:ObjectProperty} $\,\cup\,$ props used with non-literal objects; remove annotation/structural
\State $\mathcal{P}_{obj} \gets$ resulting set
\State \textbf{Describe} ($S_{\text{desc}}$): fraction of $e\in\mathcal{E}$ with any standard label/definition/synonym annotation (incl. custom \texttt{owl:AnnotationProperty} and SKOS-XL literal forms)
\State \textbf{Define} ($S_{\text{def}}$):
\Statex \quad Get $L_e$ (preferred label) and $D_e$ (concatenated definitions/comments); adequacy heuristic on $D_e$
\Statex \quad Batch BERT embeddings for $\{L_e,D_e\}$ on available device; cosine similarity $\to$ batch-normalized
\Statex \quad Per-entity score $0.4\times\text{similarity}+0.6\times\text{adequacy}$; mean $\to S_{\text{def}}$ (absent $D_e\Rightarrow 0$)
\State \textbf{Hierarchy} ($S_{\text{hier}}$):
\Statex \quad Build parent$\to$child from \texttt{subClassOf}, restriction fillers, \texttt{equivalentClass}/\texttt{intersectionOf}; drop self-edges
\Statex \quad Max depth and mean branching; normalize to targets (depth 5, breadth 3); average and round to $0\ldots10$
\State \textbf{Connection} ($S_{\text{conn}}$):
\Statex \quad For $e\in\mathcal{E}$, accumulate distinct $\mathcal{P}_{obj}$ used by $e$ (direct and via restrictions) and total incident links
\Statex \quad Compute coverage, diversity (distinct props capped at 5), richness (log-scaled degree); weighted sum $\to 0\ldots10$
\State \textbf{return} $\{S_{\text{desc}},S_{\text{def}},S_{\text{conn}},S_{\text{hier}}\}$
\end{algorithmic}
\end{algorithm}

\subsection{Well-Defined Metric}

The \textbf{Well-Defined} metric assesses the semantic quality and utility of an entity's textual definitions. 
Building upon the foundational existence check, this score quantifies whether a definition is both 
semantically relevant to the entity’s label and textually adequate as a human-readable explanation.  
Each entity’s score is a weighted combination of two distinct sub-metrics: 
\textit{Semantic Relevance} (40\% weight) and \textit{Textual Adequacy} (60\% weight).

\subsubsection*{1. Semantic Relevance (40\% Weight)}
This sub-metric measures the semantic relationship between an entity's label and its definition using a Bidirectional Encoder Representations from Transformers (BERT) model \cite{devlin2019bert}.

\paragraph{Text Collection.}
For each entity, the system extracts its primary label—prioritizing preferred labels such as \texttt{skos:prefLabel} before falling back to \texttt{rdfs:label}. 
Separately, it concatenates all available textual definitions from properties such as 
\texttt{rdfs:comment}, \texttt{skos:definition}, and other common definition predicates.

\paragraph{Embedding and Similarity.}
The BERT model generates embeddings for both the label and the concatenated definition text. 
A cosine similarity is then computed between these embeddings. 
The raw similarity values are normalized using a sigmoid transformation, based on the mean 
and standard deviation of all similarity scores in the batch. 
This normalization yields a stable semantic relevance score ranging from $0$ to $1$.

\subsubsection*{2. Textual Adequacy (60\% Weight)}
This sub-metric evaluates the quality of the definition text itself, penalizing definitions that 
are trivial, incomplete, or uninformative.

\paragraph{Completeness}
The total number of tokens in each definition is measured and normalized 
against a minimum target length, ensuring that definitions are sufficiently detailed.

\paragraph{Quality}
A linguistic heuristic calculates the ratio of non-stop-words 
(meaningful tokens) to the total number of tokens, promoting information-rich content.

\subsubsection*{Score Normalization}
For each entity $e$, the composite score is calculated as:
\[
\text{Score}_e = (0.4 \times \text{Relevance}_e) + (0.6 \times \text{Adequacy}_e)
\]
If an entity lacks any definition text, $\text{Score}_e = 0$.  
The final \textbf{Well-Defined Metric} is computed as the mean of all entity scores, scaled to a 10-point scale:
\[
\text{Well-Defined Score} = 10 \times \frac{\sum_{i=1}^{N} \text{Score}_i}{N}
\]
where $N$ is the total number of entities in the ontology.

\subsection{Connection Metric}

The \textbf{Connection} metric evaluates the structural interlinking of entities within an ontology. 
It quantifies how effectively the ontology captures semantic relationships between entities, 
beyond mere labeling or annotation.

\subsubsection*{Predicate Set Identification}
The calculation first identifies the set of predicates that represent semantic connections. 
This set primarily includes properties explicitly declared as \texttt{owl:ObjectProperty}, as well as 
other properties inferred from usage to link entities (i.e., their object is not a literal).  

To ensure that the metric focuses on genuine structural links, the system explicitly excludes:
\begin{itemize}
\item Annotation properties such as \texttt{rdfs:label} and \texttt{skos:definition}.
\item Graph-structural predicates such as \texttt{rdf:type} and \texttt{rdfs:subClassOf}.
\end{itemize}

\subsubsection*{Metric Calculation}
The system iterates through all entities—both classes and individuals—examining their direct triples as 
well as relationships defined within \texttt{owl:Restriction} axioms. For each entity, connection data are 
aggregated to compute three sub-metrics, which together form the overall Connection Score.

\paragraph{Coverage (70\% Weight)}
Measures the proportion of all entities that participate in at least one connecting relationship.  
This provides a baseline indicator of the ontology’s overall connectedness.

\paragraph{Diversity (20\% Weight)}
Calculates the average number of distinct connecting predicates associated with each entity, normalized 
against a target value of 5. This rewards ontologies in which entities exhibit a variety of relationships, 
rather than repetitive or monolithic linkage patterns.

\paragraph{Richness (10\% Weight)}
Evaluates the average total number of connections (incoming and outgoing) per entity.  
The raw count is log-scaled and normalized (against a target of 10) to reward well-connected hub entities 
while preventing outliers from disproportionately influencing the overall score.

\subsubsection*{Score Normalization}
The three weighted sub-scores are combined and scaled to a 10-point range:

\begin{equation}
\begin{aligned}
\mathrm{ConnScore}=10\big(&0.7\,\mathrm{Coverage}+0.2\,\mathrm{Diversity} \\
                         &+0.1\,\mathrm{Richness}\big)
\end{aligned}
\label{eq:conn}
\end{equation}

This final value reflects the ontology’s balance among breadth of coverage, 
variety of relationship types, and depth of entity interconnections.

\begin{align*}
\mathcal{E} &= \text{entities (classes)} \cup \text{individuals} \\
\text{Cov} &= \frac{|\{e \in \mathcal{E} \mid \text{connections}(e) \geq 1\}|}{|\mathcal{E}|} \\
\text{Div} &= \frac{1}{|\mathcal{E}|}\sum_{e \in \mathcal{E}} \min\left(\frac{|\text{distinct\_predicates}(e)|}{5}, 1.0\right) \\
\text{Rich} &= \frac{1}{|\mathcal{E}|}\sum_{e \in \mathcal{E}} \min\left( \log_{11}(\text{total\_connections}(e) + 1), 1.0\right) \\
\text{Score} &= 10 \times \left( 0.7 \times \text{Cov} + 0.2 \times \text{Div} + 0.1 \times \text{Rich} \right)
\end{align*}

\subsection{Hierarchical Breadth Metric}

The \textbf{Hierarchical Breadth} metric evaluates the structural quality of the ontology’s class taxonomy 
by analyzing both its depth and breadth. This metric rewards hierarchies that are well-balanced—deep enough 
to be expressive, but not so deep as to become unwieldy, and with a branching factor that maintains 
conceptual clarity.

\subsubsection*{Hierarchy Construction}
A distinguishing feature of this metric is its ability to construct a comprehensive parent–child graph 
that extends beyond simple \texttt{rdfs:subClassOf} assertions.  
The system infers hierarchical relationships by traversing complex OWL axioms, including:
\begin{itemize}
\item Extracting parent–child links from fillers of \texttt{owl:Restriction} axioms 
  (e.g., \texttt{owl:someValuesFrom}).
\item Parsing \texttt{owl:intersectionOf} lists and treating the component classes as parents.
\item Analyzing \texttt{owl:equivalentClass} expressions to infer implicit parentage 
  from restrictions or intersections.
\end{itemize}
This process yields a more accurate and semantically complete representation of the ontology’s logical hierarchy.

\subsubsection*{Depth and Breadth Calculation}
\paragraph{Depth}
The system computes the maximum depth of the inferred hierarchy using an iterative, cycle-safe traversal algorithm \cite{kozen1992depth}.  
The resulting raw depth is normalized against a target maximum depth of 5 \cite{noy2001ontology}, considered an optimal balance between 
expressiveness and interpretability \cite{bergman2010executive}.    

\paragraph{Breadth}
The average number of direct child classes per parent is calculated to determine the ontology’s branching factor.  
This mean breadth value is normalized against a target of 3, consistent with best practices for manageable 
hierarchical modeling.

\subsubsection*{Score Normalization}
The final score is computed as the integer-rounded mean of the normalized Depth and normalized Breadth scores:

\begin{equation}
\begin{aligned}
\mathrm{HBScore}=\operatorname{round}\Big(10\,\tfrac{\mathrm{Depth}_{\mathrm{norm}}}{2}
\;+\;10\,\tfrac{\mathrm{Breadth}_{\mathrm{norm}}}{2}\Big)
\end{aligned}
\label{eq:hb}
\end{equation}

The result is a value between 0 and 10, representing the ontology’s overall balance between hierarchical Depth and structural Breadth.


\section{Implementation: User Interface and Visualizations }
WiseOWL offers a web application accessible via Streamlit \cite{streamlit}, a free Python library, enabling users to upload OWL files through an intuitive front end. For each metric score, a donut chart (e.g., Figs.\ref{fig:wiseowl-po},\ref{fig:wiseowl-go}) is generated using Plotly \cite{plotly}. Additionally, the interface includes an option to download data in CSV format, allowing for offline analysis and easy integration with various tools.


\section{Evaluation}

We applied our method to multiple publicly available ontologies to illustrate WiseOWL’s capabilities. This diverse set included the Plant Ontology (PO) \cite{jaiswal2005plant}, Gene Ontology (GO) \cite{ashburner2000gene}, Semanticscience Integrated Ontology (SIO) \cite{sio_bioportal}, Food Ontology (FoodON) \cite{dooley2018foodon}, Dublin Core (DC) \cite{kakali2007integrating}, and GoodRelations \cite{hepp2008goodrelations}. To demonstrate our application's analytical process, we will focus on the results from Plant Ontology (PlantON) and Gene Ontology (GO) as representative examples.

\begin{figure}[htbp]
  \centering
  \includegraphics[width=0.45\textwidth]{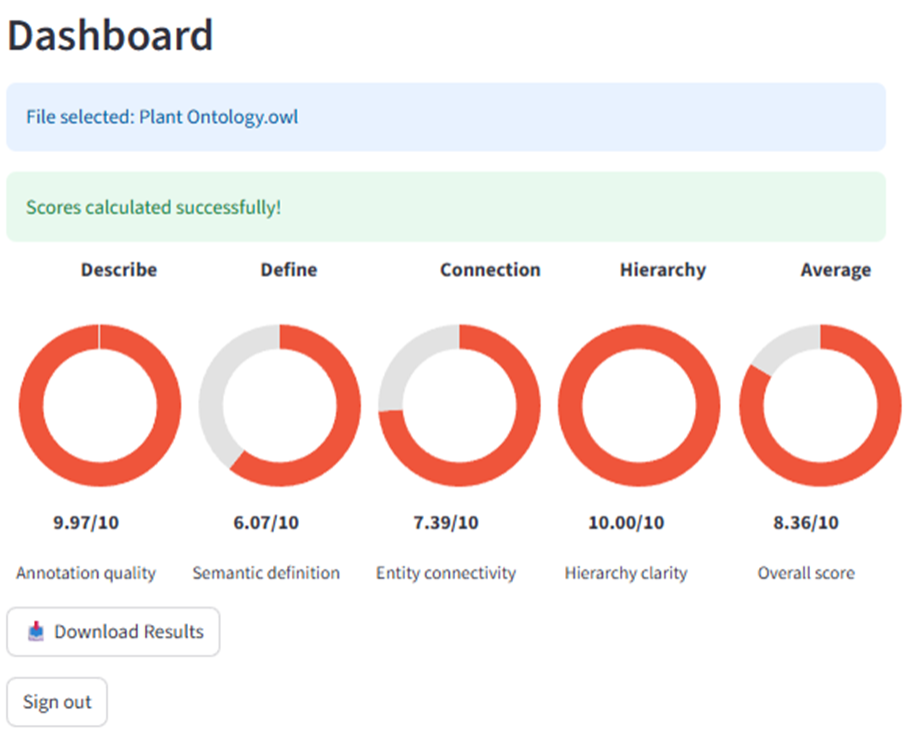}
  \caption{WiseOWL UI application displaying evaluation results for the Plant Ontology (PlantON), highlighting evaluation metrics i.e. Well-Described score, Well-Defined score, Connection score, and Hierarchical Breadth score, along with the overall average score. (Please note each metric is abbreviated as Describe, Define, Connection, and Hierarchy in the Application)}
  \label{fig:wiseowl-po}
\end{figure}

\begin{figure}[htbp]
  \centering
  \includegraphics[width=0.45\textwidth]{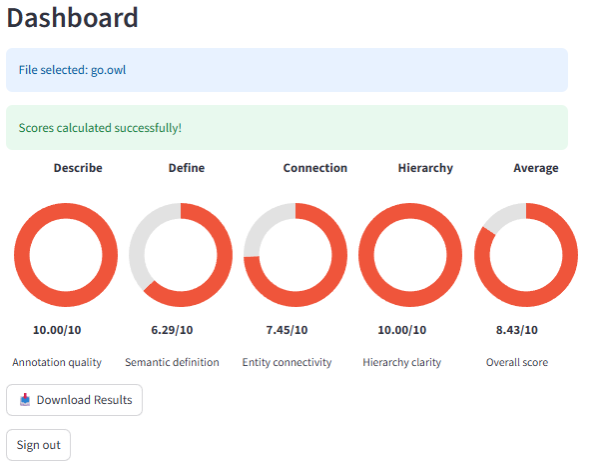}
  \caption{WiseOWL UI application displaying evaluation results for the Gene Ontology (GO), highlighting evaluation metrics -- Well-Described score, Well-Defined score, Connection score, and Hierarchical Breadth score —-along with the overall average score. (Please note each metric is abbreviated as Describe, Define, Connection, and Hierarchy in the Application)}
  \label{fig:wiseowl-go}
\end{figure}

The Plant Ontology (PO) serves as a foundational resource supporting research in genetics, genomics, phenomics, developmental biology, taxonomy and systematics, semantic applications, and education \cite{wikipedia_plant_ontology}. Its structured representation of plant entities and relationships, along with comprehensive domain coverage, makes it well suited for inclusion in WiseOWL’s evaluation to demonstrate the system’s applicability to large-scale, domain-specific ontologies. Moreover, the Gene Ontology (GO) is a dynamic \cite{go_ontology_documentation}, community-driven ontology that plays a central role in advancing biomedical research. It has been cited in tens of thousands of scientific publications for its use in interpreting large-scale molecular biology experiments and elucidating the structure, function, and dynamics of living organisms \cite{go_introduction}. Recognizing its significance in biomedical knowledge representation, the Gene Ontology (GO) is employed to evaluate WiseOWL’s performance.

As depicted in Fig.\ref{fig:wiseowl-po}, the evaluation of the Plant Ontology produced a distinct quality profile. It achieved near-perfect scores for Well-Described ('Describe' value of 9.97) and Hierarchical Breadth ('Hierarchy' value of 10.00), indicating that the ontology is well documented and structurally organized. The Well-Defined score ('Define' value of 6.07) reflects moderate adequacy of semantic definitions, while the Connection score (7.39) suggests a reasonably well-connected set of entities. The overall average score of 8.36 demonstrates WiseOWL’s capability to quantitatively capture multiple dimensions of ontology quality and highlight specific areas for refinement. Similarly, the WiseOWL assessment of Gene Ontology (GO) shown in Fig.\ref{fig:wiseowl-go} demonstrates a well-balanced internal structure with consistently high performance across WiseOWL’s evaluation metrics. The ontology achieved maximum scores for Well-Described ('Describe' value of 10.00) and Hierarchical Breadth ('Hierarchy' value of 10.00), indicating extensive documentation and a logically coherent class organization. The Well-Defined ('Define' value of 6.29) metric reflects reasonably detailed semantic definitions, while the Connection (7.45) score captures its strong relational structure among entities. The overall average score of 8.43 highlights the ontology’s maturity and semantic integrity, showcasing WiseOWL’s effectiveness in capturing ontology quality through automated, interpretable visual analysis.

\subsubsection{Runtime and Hardware}
 On an Apple M3 Pro (18\,GB RAM), WiseOWL processed $\sim$2787853 triples in $\sim$2 minutes end-to-end (parsing+scoring; BERT batch size 64, max\_length 128).

\begin{table}[!t]
\caption{Ontology scores across metrics (0–10). Higher is better.}
\label{tab:ontology-scores}
\centering
\small
\setlength{\tabcolsep}{4pt}
\renewcommand{\arraystretch}{1.1}
\begin{tabular}{lccccc}
\toprule
\textbf{Ontology} & \textbf{Describe} & \textbf{Define} & \textbf{Conn.} & \textbf{Hier.} & \textbf{Avg.} \\
\midrule
PO           &  9.97 & 6.07 & 7.39 & 10.00 & 8.36 \\
GO           & 10.00 & 6.29 & 7.45 & 10.00 & 8.43 \\
SIO          & 10.00 & 5.72 & 2.21 & 10.00 & 6.98 \\
FoodON       & 10.00 & 5.13 & 2.48 & 10.00 & 6.90 \\
DC           &  9.17 & 5.71 & 0.00 &  6.00 & 5.22 \\
GoodRelations&  9.29 & 6.09 & 3.87 &  8.00 & 6.81 \\

\bottomrule
\end{tabular}
\end{table}


\section{Limitations and Future Work}
Our preliminary evaluation of the WiseOWL ontology assessment strategy highlights four key structural properties of ontologies: Descriptive richness (Well-Described metric), semantic coherence (Well-Defined metric), interconnectedness (Connection metric), and hierarchical balance (Hierarchical Breadth metric). While these metrics provide a robust, quantitative baseline, the current implementation has limitations that define our future work.
A primary limitation is that the metrics rely on static, "one-size-fits-all" thresholds. For instance, the Hierarchical Breadth score assumes a target depth of 5 and breadth of 3, while the Connection score uses a fixed target for relationship diversity. These fixed values may not be appropriate for all domains; e.g., a complex biomedical ontology and a simple vocabulary will have different structural expectations. A key future direction is to make these thresholds user-configurable or, more advanced, to derive "gold standard" profiles by analyzing a corpus of high-quality ontologies from different domains.

A second limitation is in the Well-Defined metric, which currently uses a general-purpose model (bert-base-uncased) and a simple textual heuristic (the\_adequacy\_score). This approach is a proxy for semantic quality, but lacks true contextual understanding. As a major future direction, we aim to replace this component by incorporating semantic alignment with domain-specific corpora. Integrating large language models (LLMs) or domain-specific models (e.g., BioBERT) \cite{lee2020biobert} would enable more context-aware and meaningful evaluations of definition quality, moving from a simple heuristic to a true semantic assessment.

Finally, we plan to extend WiseOWL’s functionality to bridge the gap between its quantitative scores and qualitative, actionable advice. This includes recommendations for relevant Ontology Design Patterns (ODPs) \cite{gangemi2009ontology} based on the metric analysis. For instance, a low Connection score may suggest the use of the Participation or Agent-Role pattern to enhance relational depth, while a low Hierarchical Breadth score could indicate the need for refactoring using n-ary Relation or Normalization patterns to improve structural clarity.

\section{Conclusion }
The WiseOWL methodology provides a quantitative framework for measuring ontologies along the dimensions of descriptive richness, semantic coherence, interconnectedness, and hierarchical balance. The scores that it assigns are derived from the assessed ontology’s schema, which serves as the foundational criterion for evaluation. Using existing ontologies for testing allows us to observe how this application generates relevant and impactful suggestions for ontology improvement. Integrating WiseOWL with large language models (LLMs) could further enhance these results by enabling more context-aware and intelligent assessments of ontologies. 

\bibliographystyle{IEEEtran}
\bibliography{references}

\end{document}